\documentclass[letterpaper, 10 pt, conference]{ieeeconf}  % Comment this line out if you need a4paper

\usepackage{amsmath}
\usepackage{xcolor}
\usepackage{mdframed} 
\usepackage{amsthm}

\usepackage{subcaption}  
\usepackage{siunitx}          
\usepackage{amsfonts}
\usepackage{balance} %
\usepackage{lipsum} 
\usepackage{float}
\usepackage{graphicx} 
\usepackage{booktabs}
\usepackage{url}
\usepackage{amssymb}
\usepackage{multirow}  
\usepackage[ruled,vlined]{algorithm2e} 
\usepackage{cite}
\usepackage{multirow} 
\usepackage{mdframed}          
\usepackage{algpseudocode}      % npseudocode commands
\IEEEoverridecommandlockouts                 
\title{\LARGE \bf
ContactRL: Safe Reinforcement Learning based Motion Planning for Contact based Human Robot Collaboration}

\author{
        Sundas Rafat Mulkana,
        Ronyu Yu,
        Tanaya Guha,
        Emma Li\\
\textit{School of Computing Science, University of Glasgow}}

\begin{document}
\maketitle
\thispagestyle{empty}
\pagestyle{empty}

\begin{abstract}
In collaborative human-robot tasks, safety requires not only avoiding collisions but also ensuring safe, intentional physical contact. We present ContactRL, a reinforcement learning (RL) based framework that directly incorporates contact safety into the reward function through force feedback. This enables a robot to learn adaptive motion profiles that minimize human-robot contact forces while maintaining task efficiency. In simulation, ContactRL achieves a low safety violation rate of 0.2\% with a high task success rate of 87.7\%, outperforming state-of-the-art constrained RL baselines. In order to guarantee deployment safety, we augment the learned policy with a kinetic energy based Control Barrier Function (eCBF) shield. Real-world experiments on an UR3e robotic platform performing small object handovers from a human hand across 360 trials confirm safe contact, with measured normal forces consistently below 10N. These results demonstrate that ContactRL enables safe and efficient physical collaboration, thereby advancing the deployment of collaborative robots in contact-rich tasks. \textit{Implementation is available at \url{https://github.com/SMulkana/ContactRL}}.

\end{abstract}


\section{INTRODUCTION}
In the era of Industry 5.0, human–robot collaboration increasingly involves physical contact \cite{zafar2024exploring}. In precision manufacturing and related tasks, robots may need to retrieve small tools such as screws  (or even delicate objects like gems in jewelry workshops) directly from a human hand \cite{costanzo2021handover}. The palm-up orientation is considered the most secure for such handovers \cite{duan2024human}. However, attempting to grasp objects this way can lead to unintended contact with the hand. This raises the central question our work addresses: \textit{how can robots pick up small objects from a human hand without causing discomfort or injury}? We study this task as a representative case where safe, unavoidable contact is essential for task execution. Here, safety is defined in terms of limiting the contact normal force, $F_N$, exerted on the human hand by the robot during physical contact (see Fig. 1).

\begin{figure}
    \centering
    \includegraphics[width=0.9\linewidth]{Figure1-Problem.png}
    \caption{Positional-deviation profile $\mathbf d(t)$ from start $t_0$ to finish $t_N$. We seek a trajectory minimizing task completion time that satisfies the terminal safety constraint $F_N(\mathbf d(t_N))\le F_\tau$, where $F_N$ is the contact force on the hand and $F_\tau$ is the safe contact limit, to ensure human comfort and safety during small object handover.}
    \label{fig:problem}
    \vspace{-0.3cm}
\end{figure}

Reinforcement Learning (RL) has shown success in manipulation tasks such as pick-and-place and object handover\cite{lobbezoo2021reinforcement}. Recent work in RL based motion planning has increasingly focused on safety, with particular attention to collision avoidance in static and dynamic environments \cite{hu2023adaptive}. Most safe motion planning approaches treat contact as a failure event to be avoided. Yet in many collaborative tasks, contact is necessary for task execution (e.g., handovers or assistive tasks). Ensuring low contact force while achieving task goals such as rapid and precise reaching introduces complex trade-off between safety and performance which are not addressed by conventional collision avoidance frameworks \cite{yu2024pathrl}.

To address this gap, we propose ContactRL, a reinforcement learning framework that explicitly integrates contact safety, quantified via contact force, as a reward term alongside task-relevant criteria such as reaching and motion smoothness. This formulation enables the learning of motion policies that actively balance task performance with low-force interaction. However, RL methods lack formal safety guarantees, which is a critical requirement for robots collaborating closely with humans. To address this limitation, we integrate a kinetic energy based Control Barrier Function (eCBF) safety shield \cite{choudhary2022energy} at deployment. Unlike prior safe-RL methods, which equate safety with non-contact, our work addresses the underexplored but critical scenario where contact is intentional and unavoidable. In simulation, ContactRL has a 0.2 \% rate of safety-constraint violations. On the physical robotic platform, ContactRL employed with the eCBF shield was tested across 360 trials with 12 participants. The contact forces (measured with UR3e force/torque sensor) remained consistently below 10\text{N}, which is within the safe human contact limit of 50\text{N} \cite{bilyea2023modeling}.

\subsection{Related Works}

\subsubsection{\textbf{Force modulation}} 
Compliance control is widely adopted to achieve safe contact; this includes techniques such as impedance control and admittance control. However, these techniques heavily rely on precise environmental models, making them less adaptable to dynamic and unstructured environments particularly those involving humans \cite{dani2020human}. Hybrid compliance approaches such as hybrid Impedance–Admittance \cite{ye2024hybrid} and Variable‑Impedance \cite{martin2019variable} while improve tracking performance and adaptability, rely on labor‑intensive gain‑tuning (e.g., stiffness/damping) to keep contact forces within safe limits, making it difficult to guarantee safety–efficiency trade‑offs across tasks. Learning from demonstration, integrated with weighted random forests, has been explored for force modulation through path generation \cite{al2021improving}. However, this approach failed to maintain forces within the prescribed threshold. RL, on the other hand, does not rely on explicit environmental models making it suitable for dynamic environments \cite{melnik2021using}. 
\subsubsection{\textbf{Safe RL based robot motion planning}} There are three main methodologies for enforcing safety in RL-based robotic tasks: (1) \textit{Reward shaping} - penalizes unsafe behaviors to guide agents toward safer actions through policy learning \cite{liu2021deep}. Recent works have explored safety-aware reward functions \cite{miranda2023generalization}, and also incorporated co-efficiency metric into the reward function \cite{lagomarsino2022robot}. The above-mentioned works focus on collision avoidance without considering safe contact. (2) \textit{Constrained RL} - where safety constraints are explicitly added to the cost functions and optimized jointly with reward functions. This includes Model Predictive Control (MPC)\cite{koptev2024reactive} and Lagrangian-based optimization \cite{sun2024force} that integrates force thresholds into the optimization objective. The work closest to ours \cite{sun2024force} employs a dynamic force prediction model for the task of assistive dressing. It ensures safe interaction by predicting forces within predefined limits. However, mediated interaction models like this one do not generalize well to tasks that require precise, direct contact. (3) \textit{Shielding} or barrier methods employ a predefined safety filter (shield) which assesses each action before execution. If an action violates safety constraints, the shield replaces it with a safer alternative, often selected from a predefined safe set or a modified version of the original action \cite{brunke2022safe}. However, the shield's action replacement can cause the robot to take abrupt actions, leading to jerky movements \cite{pham2018optlayer}. This affects the fluidity of interaction and makes it less suitable for safe contact scenarios. 
\subsubsection{\textbf{Safe RL via barrier based methods}}
Recent research efforts have focused on enabling analytical safety guarantees for RL based motion policies through runtime monitoring frameworks. Control Barrier Functions (CBFs) are widely used to enforce safety constraints analytically during policy execution \cite{guerrier2024learning}. For instance, the CRABS algorithm integrates the learning of barrier functions, system models, and policies to prevent safety violations; however, its reliance on learned models and pre-trained certificates makes it less suited for scenarios with complex contact dynamics \cite{luo2021learning}. Similarly, disturbance observer based CBFs require reliable dynamics estimation, which is often unavailable or inaccurate in RL-generated trajectories \cite{cheng2023safe}. To support both safety and task stability, hybrid approaches combining Control Lyapunov Functions (CLFs) with CBFs have been proposed to maintain stability and safety using learned control policies \cite{du2023reinforcement}. 
\subsection{Contributions}
In contrast to the past works, we propose to achieve \textit{force modulation through RL-based motion planning by incorporating force feedback based reward in policy learning}. Our approach balances safety and performance in physical human-robot interactions, contributing to the broader goal of enabling safe and effective collaborative robotics. In summary, the contributions of this work are as follows:
\begin{enumerate}
    \item We present ContactRL, a novel RL-based motion planning framework, for safe human-robot contact. Our RL policy plans a safe motion profile for UR3e robot, modulating the human-robot contact force ($F_N$). 
    \item In simulation our method commits only one safety violation out of 1000 evaluation episodes, outperforming state-of-the-art constrained RL baselines.
    \item We augment the policy with a kinetic energy based CBF shield upon deployment for guaranteed safety. 
    \item We validate the shielded policy on real world robotic platform with 12 participants across 360 physical hand‑over trials confirming $F_N$ remains below 10N.   
\end{enumerate}

\begin{figure}
    \centering
    \vspace{10pt} % Adds space before the figure
    \includegraphics[width=1\linewidth]{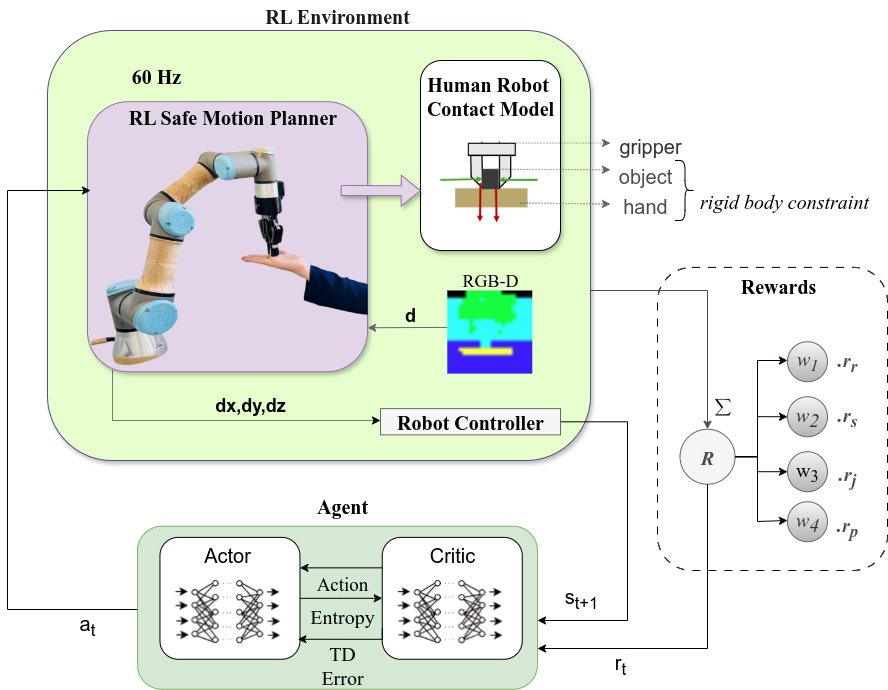}
    \caption{ContactRL: An RL-based motion planning framework for human–robot close-contact interaction. The robot learns an adaptive motion profile based on a human-robot contact model. A rigid-body constraint is applied between the robot and the object to enable lift, while the normal contact force exerted on the human hand, ($F_N$) is feedback to the safety component of the reward function.}
    \label{fig:fig:2}
    \vspace{-0.7em} % adjust value as needed
\end{figure}

\section{Contact RL}

In ContactRL, the RL agent generates a safe motion profile by optimizing its policy through receiving rewards for safe interactions and penalties for unsafe contact using force-feedback, which is shown in Fig. 2. The robot utilizes visual perception to extract the spatial coordinates of the robot end-effector and the object within the workspace. This extracted information constitutes the observation space for the RL agent, which subsequently executes the next action by taking a step toward the target object. The control architecture consists of high-level motion planning governed by the RL agent. An inverse kinematic controller translates the planned end-effector poses into joint pose, which are then executed using position control. To ensure human safety, the force exerted by the robot on human hand, $F_\text{N}$ must remain below a predefined human discomfort threshold value, $F_{\tau}$. 

\subsection{Markov Decision Process}
In our setup, the “pick-from-hand-safely’’ task is modeled as the MDP
$\langle\mathcal{S},\mathcal{A},P,R,\gamma\rangle$, where,
\subsubsection{Observation space (S)} We formulate the state space as 9-D observation vector;  
$\displaystyle 
\mathbf{S}=[\,p_{\text{r}},\,p_{\text{h}},\,v_{\text{r}}\,]$,
where $p_{\text{r}},p_{\text{h}}\!\in\!\mathbb{R}^3$ are robot end-effector and hand poses and $v_\text{r} \!\in\!\mathbb{R}^3$ is end effector velocity. Hand position is randomized at each episode.
\subsubsection{Action space (A)}  We define the policy's action space as a 3-D continuous space corresponding to Cartesian displacements of the robot's end-effector. At each control timestep, the policy outputs an action vector $\boldsymbol{a} = [\delta_x, \delta_y, \delta_z] \in \mathbb{R}^3$, which specifies a desired relative movement in the end-effector's position along the $x$, $y$, and $z$ axes. To ensure smooth and safe execution, the magnitude of each action is bounded such that $\| \boldsymbol{a} \| \leq \delta_{\text{max}}$, where $\delta_{\text{max}}$ denotes the maximum step size per dimension. 
\subsubsection{Reward design (R)} We develop a dense reward function that balances performance and safety considerations elected through empirical evaluation of multiple reward combinations. Here:

\begin{equation}
\label{eq:reward_weighted}
\begin{aligned}
R(\mathbf{s},\mathbf{a}) &=
\;w_r\,
\underbrace{
\begin{cases}
500 & \text{contact bonus}\\
-\,\lVert \mathbf{p}_r - \mathbf{p}_h \rVert & \text{otherwise}
\end{cases}}_{\text{reach reward, } r_r}
\\[6pt] &\quad
+\,w_s\,
\underbrace{
\begin{cases}
0 & \text{no contact}\\
F_{\tau}\left(1 - \frac{F_N}{F_{\tau}}\right) & \text{contact and } F_N \le F_{\tau}\\
-\;F_{\tau}\,\frac{F_N - F_{\tau}}{F_{\tau}} & \text{contact and } F_N > F_{\tau}
\end{cases}}_{\text{safety reward, } r_s}
\\[6pt] &\quad
+\,w_j\,
\underbrace{\bigl\lvert \delta^{t} - \delta^{t-1} \bigr\rvert}_{\text{jerk reward, } r_j}
\;+\;
w_p\,
\underbrace{\left(\frac{\lVert \mathbf{p}_r - \mathbf{p}_h \rVert}{\lVert \mathbf{p}_h \rVert}\right)
           \lVert \delta \mathbf{p}_r \rVert}_{\text{proximity reward,
           } r_p}
\end{aligned}
\end{equation}
here $w_r,w_s,w_j,$ and $w_p$ are weights for reach, safety, jerk and proximity rewards (as shown in Table I) which are varied in an ablation study to select the optimal reward weighting; $\mathbf{p}_{r}$ and $\mathbf{p}_{h}$ are robot and hand positions at each iteration; and $\delta^{t}$ and $\delta^{t-1}$ are current and previous step size.

Reach reward encourages the agent to move end-effector towards the hand by penalizing the distance between the hand and robot. In addition, we give a contact bonus (500) when the end effector reaches the object. Safety reward incentivizes maintaining contact forces below a threshold $F_\tau$ by giving a positive reward for gentle contact and applying a penalty for excessive force. Jerk reward penalizes large differences between consecutive actions to minimize variation between two consecutive actions. Proximity reward encourage the robot to take larger steps at the start and smaller as it approaches the human to minimize completion time. Through this reward design, the robot is encouraged to modulate it's step size to ensure safe contact with human hand.
\subsubsection{Transition dynamics (P)} 
The $P(\mathbf{s}'\!\mid\!\mathbf{s},\mathbf{a})$ follows deterministic
rigid-body dynamics in PyBullet \cite{mower2023ros} with 1/60~\text{Hz} simulation frequency.
An episode terminates upon robot grasping object or truncates when episodes progresses to 150 timesteps. The discount factor ($\gamma$) is 0.99.
\subsection{Simulation Environment}
The policies are trained on a UR3e robotic manipulator with Robotiq 2F-85 gripper in a PyBullet \cite{mower2023ros} simulated environment. Modeling an entire human body (which comprises of complex shapes) in simulation, increases the computation time and complicates training process \cite{fang2023human}. Thus, we simplify the human hand to a flat plane model \cite{yu2021human} . In the proposed task, the contact force $F_N$, depends on the friction coefficient and the mass of the human hand. Therefore, as long as these physical properties are accurately defined, the simulation can closely mimic real-world hand physics. In simulated environment, the hand is the parent body with 0.5 kg mass (average mass of human hand)  and lateral friction value of 1 and the object is the child body with 0.002 kg and the lateral friction coefficient taken as 0.5 \cite{herbster2023modeling}. The gripper is kept
slightly open ($0.21$ rad) until contact. We keep the safe contact force threshold value, $F_\tau$, as 50~\text{N} \cite{herbster2023modeling} exceeding which will cause discomfort to human.

\begin{figure}
    \centering
    \vspace{5pt}
    \includegraphics[width=1\linewidth]{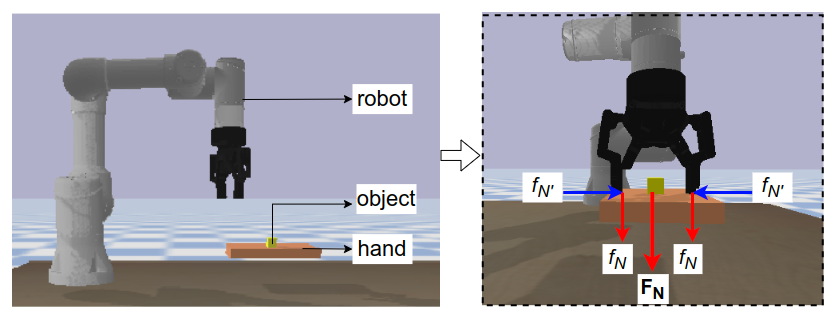}
    \caption{Simulated environment with UR3e robot, object, and a simplified hand model. The right panel illustrates the components of the contact force applied to the hand during interaction.}
    \label{fig:placeholder}
\end{figure}
\subsection{Training Setup}
We train our policy using the \textit{Soft Actor–Critic} (SAC) algorithm, an off-policy method for continuous action spaces. SAC optimizes a stochastic policy to maximize both expected return and entropy, promoting exploration and stability. We use the continuous-action variant of SAC from the \texttt{Stable-Baselines3} library~\cite{raffin2021stable}, implemented in PyTorch. We retain the library’s default hyperparameters: a learning rate of $3 \times 10^{-4}$ for both actor and critic, a replay buffer of $10 \times 10^{6}$ transitions with FIFO sampling, a batch size of 256, and soft target updates with $\tau = 0.005$. One gradient update is performed per environment step, and the entropy coefficient is automatically tuned. Both policy and value networks are two-layer feedforward models with 256 hidden units per layer and ReLU activations.  

For robustness, we train with five independent seeds (\texttt{0}--\texttt{4}), each controlling all stochastic elements such as environment initializations, action sampling, weight initialization, and buffer sampling. Episodes are capped at a horizon of 150 timesteps. For each seed we conduct a separate training run with identical settings, and report the mean and standard deviation of episodic returns across the five runs. 

\subsection{Energy Barrier Safety Shield}\label{sec:method}

To ensure safety during policy execution, we adapt the eCBF shield \cite{choudhary2022energy} to regulate contact force implicitly via end-effector kinetic energy. Unlike force model based CBFs, which require precise modeling of contact dynamics, the kinetic energy based approach defines safety in terms of instantaneous task-space energy. Our goal is to guarantee (i) that the end–effector never exceeds the 50~\text{N} contact force limit and (ii) that it still converges to the desired grasp pose.  We achieve this by solving a single Quadratic Program (QP) that unifies an energy–based zeroing CBF with a goal–tracking CLF. This allows us to embed real-time force safety into policy execution, acting as a safety filter that dynamically corrects the commanded motion without overriding the learned policy. 

The shield works in two phases (Algorithm 1): \textit{a) intervention} and \textit{b) action replacement}. At each control cycle, the SAC policy outputs a Cartesian displacement $\Delta\mathbf{p}\in\mathbb{R}^{3}$, which is converted into a raw Cartesian velocity command $\mathbf{v}_{\text{raw}}\in\mathbb{R}^{3}$ for the low-level controller. We first smooth this signal with a first-order low-pass filter (LPF) to suppress the $100$--$200\,$\si{\hertz} jitter.  The filter is implemented recursively as
    $\mathbf v_{f}(k)
    = \alpha\,\mathbf v_{f}(k-1) + \bigl(1-\alpha\bigr)\,\mathbf v_{\text{raw}}(k)$,
where $\alpha = e^{-2\pi f_c\Delta t}$; $f_c=25\,$\si{\hertz} is the cut-off frequency and $\Delta t=4\,$\si{\milli\second} is the control period.  This preserves the low-frequency approach motion while attenuating sub-frame impulses by more than \SI{10}{dB}.  The smoothed command $\mathbf v_{f}$ is then passed to a eCBF that enforces the safety invariant $\frac{1}{2}m\|v\|^2 \le E_{\text{max}}$ with $m=\SI{0.93}{\kilogram}$ (UR3e\,{+}\,Robotiq 2F-85) and a empirically chosen budget $E_{\max}=0.30$\,\si{\joule} based on safe force limits.  Defining $h(\mathbf x)=E_{\max}-\tfrac12 m\lVert\dot{\mathbf p}\rVert^{2}$, Nagumo’s forward-invariance condition gives the linear constraint
$-m\,\dot{\mathbf p}^{\!\top}\mathbf u + \alpha_h h(\mathbf x) \ge 0$.
The resulting QP admits a closed-form projector: if the commanded speed exceeds
$\bigl(2E_{\max}/m\bigr)^{1/2}$ the vector is radially scaled, otherwise it passes unchanged.

The proposed eCBF is a lightweight wrapper inserted between the RL policy and the robot driver. It adds only a few vector operations per tick, needs no retraining, and its correctness is provable independently of policy. The shield guarantees safety by enforcing forward-invariance of a bounded energy set, defined as $\frac{1}{2}m\|v\|^2 \le E_{\text{max}}$. Since the contact force magnitude $F_N$ scales with the rate of change of momentum, this energy constraint implicitly bounds the maximum achievable contact force, ensuring $F_N \le F_{\tau}$ during interaction. The eCBF shield achieves this by projecting the commanded velocity or control input onto the closest admissible set that satisfies this energy constraint, thereby guaranteeing safe and dissipative contact behavior.

\section{Experiments}
\subsection{Ablation Study}

We conducted an ablation study by varying the reward weights for reach, safety, jerk, and proximity to find optimal reward weighting. As shown in Table I, for RF1, we set $w_r=1$ and the rest of reward weights set to $0$. This means that we consider reach-only ablation for RF1, where the policy attains the highest cumulative reward but does so at the expense of safety, producing frequent impulsive contacts that exceed the F = 50~\text{N} threshold in 52\% of the roll-outs, as shown in Fig. 4(b). Adding a force-safety reward in RF2 immediately reduces the violation rate to 13\% while leaving both reward and completion time unchanged. Introducing a jerk regularizer $w_j$ = 1 in RF4,  tightens speed distribution, yet the absence of any proximity term means the controller can still approach the target too brusquely. RF5, which combines reach accuracy with safety, jerk, and proximity reward ($w_p$ = 1), resolves the remaining trade-off, achieving the fastest safe completion time, a median peak force below 10 N, and a residual violation rate of only 0.2\%. Fig. 4 shows that policy behavior is sensitive to the balance between safety and task performance. Excessive weighting on safety (e.g. RF3, where $w_s = 2$) caused overly conservative motions, while neglecting jerk or proximity penalties led to abrupt approach behavior. The chosen RF5 configuration emerged as the most balanced, delivering both high success rate and minimal safety violations.

\begin{table}
\centering
\vspace{10pt}
\caption{Reward function ablation variants with different weights of reward components: reach reward ($w_r$), safety reward ($w_s$), jerk reward ($w_j$), and proximity reward ($w_p$). These weights (RF1–RF5) are used to train models for the ablation study in Fig. 4.}
\begin{tabular}{lccccc}
\toprule
Variant & $w_r$ & $w_s$ & $w_j$ & $w_p$ \\
\midrule
RF1        & 1 & 0 & 0 & 0 \\
RF2              & 1 & 1 & 0 & 0 \\
RF3              & 1 & 2 & 0 & 0 \\
RF4              & 1 & 2 & 1 & 0 \\
RF5              & 1 & 2 & 1 & 1 \\
\bottomrule
\end{tabular}
\label{tab:reward_variants}
\end{table}

\begin{algorithm}
\caption{ContactRL policy with eCBF shield episode rollout}
\label{alg:ke_shield}
\KwIn{Trained policy $\pi_\theta$, environment $\mathcal{E}$, step size $\Delta t$, end effector mass $m$, kinetic energy (KE) limit $E_{\text{max}}$, LPF cutoff $f_c$}
\KwOut{Traces of distance, KE, speed, and contact force}

\BlankLine
$\alpha \gets \exp(-2\pi f_c \cdot \Delta t)$ \tcp*[f]{LPF coefficient}

$s \gets \mathcal{E}.\texttt{reset}()$ \;
$a_{\text{raw}}, \_ \gets \pi_\theta(s)$ \;
$a_{\text{filt}} \gets a_{\text{raw}}$ \tcp*[f]{Initialize LPF state}

Initialize logs: $D, KE, SPD \gets [~]$ \;
$KE_{\text{prev}} \gets 0$ \tcp*[f]{Hold previous KE for contact}

\While{episode not done}{
    $a_{\text{raw}}, \_ \gets \pi_\theta(s)$ \tcp*[f]{Policy action}
    
    $a_{\text{filt}} \gets \alpha \cdot a_{\text{filt}} + (1-\alpha) \cdot a_{\text{raw}}$ \tcp*[f]{LPF}
    
    $v_{\text{curr}} \gets \texttt{GetEEVelocity}(s)$ \;
    
    $v_{\text{mag}} \gets \|a_{\text{filt}}\|$ \;
    $v_{\text{max}} \gets \sqrt{2 E_{\text{max}} / m}$ \;
    
    \eIf{$v_{\text{mag}} \leq v_{\text{max}}$}{
        $a_{\text{safe}} \gets a_{\text{filt}}$
    }{
        $a_{\text{safe}} \gets a_{\text{filt}} \cdot (v_{\text{max}} / v_{\text{mag}})$ \tcp*[f]{Apply KE barrier}
    }
    
    $s', \_, \texttt{done}, \texttt{trunc}, \_ \gets \mathcal{E}.\texttt{step}(a_{\text{safe}})$ \;
    
    $v' \gets \texttt{GetEEVelocity}(s')$ \;
    $spd \gets \|v'\|$ \;
    $ke \gets \frac{1}{2} m \cdot spd^2$ \;
    $d \gets \texttt{GetDistanceToGoal}(s')$ \;

    Append $spd$, $ke$, $d$ to $SPD$, $KE$, $D$ \;

    \If{\texttt{done} \textbf{or} \texttt{trunc}}{
        $f_{\text{contact}} \gets \texttt{GetContactForce}(\mathcal{E})$ \;
        $ke_{\text{contact}} \gets KE_{\text{prev}}$ \;
        \textbf{break}
    }
    
    $KE_{\text{prev}} \gets ke$ \;
    $s \gets s'$
}

\Return $D$, $KE$, $SPD$, $ke_{\text{contact}}$, $f_{\text{contact}}$
\end{algorithm}

\begin{figure}
    \centering
    \begin{subfigure}[t]{0.9\linewidth}
        \centering
        \includegraphics[width=\linewidth]{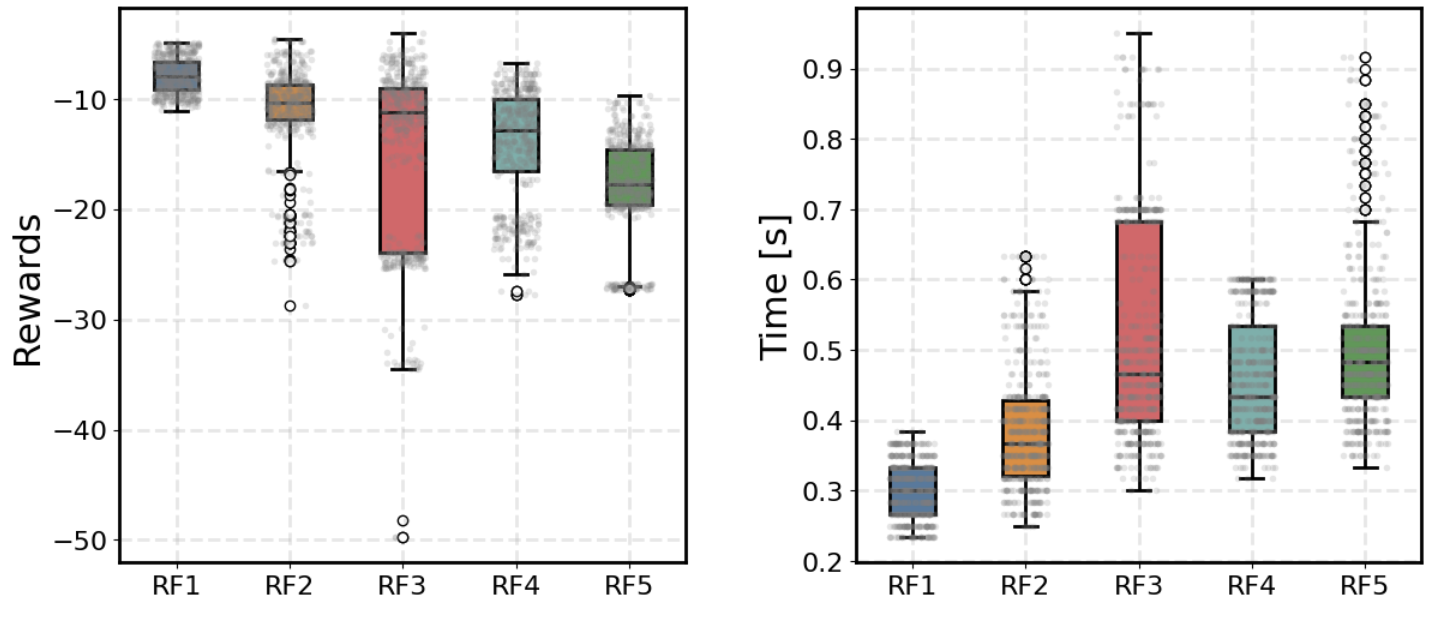}
        \caption{Performance measured by mean cumulative rewards (\textit{left}) and completion time (\textit{right}).}
        \label{fig:ablation-performance}
    \end{subfigure}
    
    \vspace{0.5em}  % Optional: small vertical space between subfigures
    
    \begin{subfigure}[t]{0.9\linewidth}
        \centering
        \includegraphics[width=\linewidth]{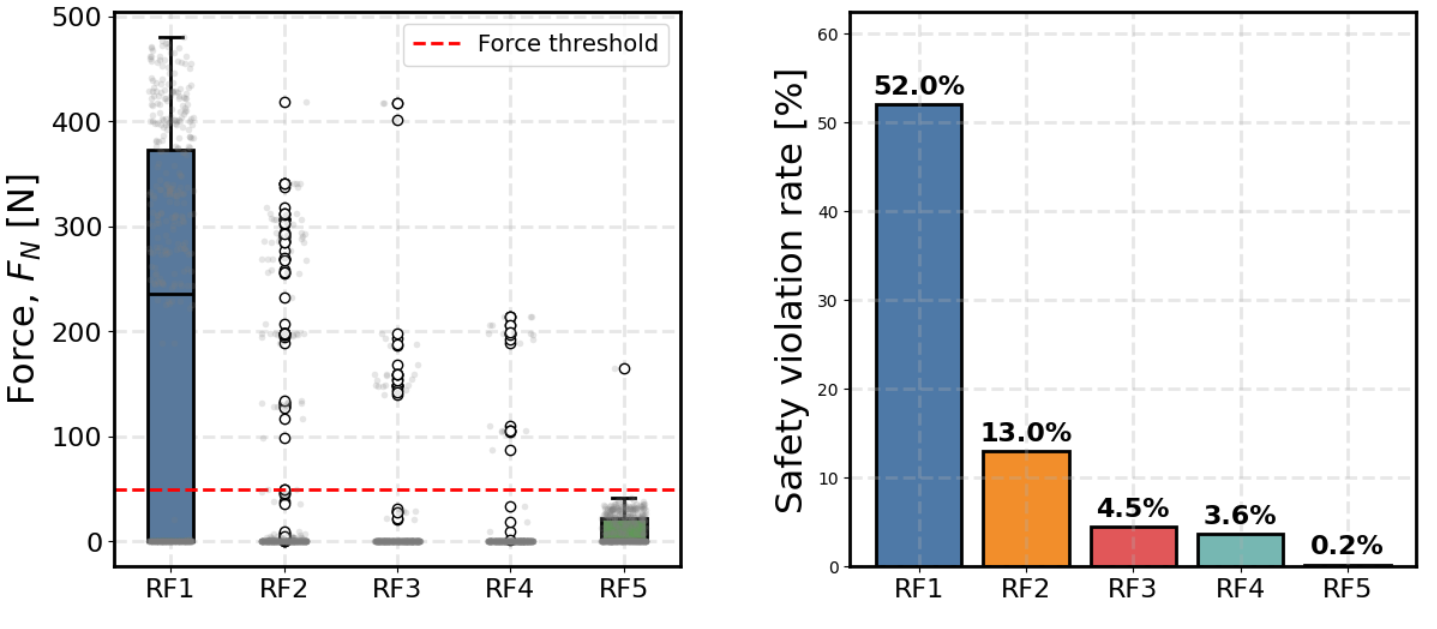}
        \caption{Safety outcomes measured by contact force (\textit{left}) relative to the safety threshold (50 $N$) and the rate of safety violations (\textit{right}).}
        \label{fig:ablation-safety}
    \end{subfigure}
    
    \caption{Ablation study of ContactRL reward function variants (RF1–RF5) with component weightings specified in Table I, evaluated over 1000 simulated episodes. RF5 achieves strong performance with minimal safety violations, outperforming other variants.}
    \label{fig:ablation-combined}
    \vspace{-0.3cm}
\end{figure}

\subsection{SOTA Comparison in Simulated Environment}

As shown in Table. II, we compare ContactRL with state of the art constrained RL baselines Constrained Policy Optimization (CPO) and SAC-Lagrangian (SACLag). We evaluate the models in simulation across 1000 episodes with randomized positions for hand in each episode. All baselines received the same observation space, including force feedback channels, to ensure comparability. The comparison metrics are success rate, time taken to perform task ($T$), mean and maximum force exerted on hand ($F_\text{N,mean}$ and $F_\text{N,max}$), number of safety violations, safety violation rate and RMS jerk. These metrics are used to compare performance, safety and motion profile. ContactRL significantly outperforms both CPO and SACLag in task success rate (87.17\% vs. 33.33\% and 55.60\%, respectively) and achieves competitive average completion times. The safety violation rate is an order of magnitude lower for ContactRL (0.20\%) compared to CPO (2.41\%) and equal to that of SACLag (0.20\%). In terms of motion smoothness, ContactRL exhibits substantially lower RMS jerk (931.32 m/s³) compared to both CPO (1355.47 m/s³) and SACLag (3924.11 m/s³). Overall, ContactRL achieves superior performance while maintaining safety and stable motion.

\begin{table*}
  \centering
  \vspace{10pt}
  \footnotesize
  \caption{Evaluation metrics for ContactRL and baseline algorithms evaluated over 5 seeds and 1000 simulated episodes. ContactRL outperforms the baselines in reaching performance, safety and motion smoothness.}
  \renewcommand{\arraystretch}{1.15}
  \setlength{\tabcolsep}{4pt}   % a bit tighter than default

  \begin{tabular}{@{}l|cc|ccc|c@{}}
    \toprule
          & \multicolumn{2}{c|}{\textit{Task Performance}} 
          & \multicolumn{3}{c|}{\textit{Safety}} 
          & \multicolumn{1}{c}{\textit{Motion Smoothness}} \\[-2pt]
    \cmidrule(lr){2-3}\cmidrule(lr){4-6}\cmidrule(lr){7-7}
          & Success [\%] & $T$ [s] 
          & $F_{N,\text{mean}}$ [N] & Violations
          &SV rate [\%] & RMS jerk [m/s$^{3}$] \\
    \midrule
    \textbf{ContactRL (ours)} & \textbf{87.17} & $\textbf{0.50}\!\pm\!\textbf{0.12}$
               & $\textbf{9.90}\!\pm\!\textbf{14.70}$ & \textbf{1}& \textbf{0.20 }
              & $\textbf{931.32}\!\pm\!\textbf{750.15}$ \\

    CPO \cite{achiam2017constrained}      & 33.33 & $0.35\!\pm\!0.14$ 
               & $6.09\!\pm\!44.02$ & 7 & 2.41 
              & $1355.47\!\pm\!164.03$  \\

    SACLag \cite{ray2019benchmarking}    & 55.60 & $1.08\!\pm\!0.55$ 
                  & 14.77  $\!\pm\!$ 29.55                & 1& 0.20 
              & $3924.11\!\pm\!2250.75$  \\
    \bottomrule
  \end{tabular}
  \label{tab:cbf_comparison_no_shield_transposed}
\end{table*}

\subsection{Shielded vs. Unshielded Policy}
In simulated evaluations, we found that with stochastic actions which represent real-world dynamics, the ContactRL policy alone cannot guarantee safety. As shown in Fig. 5(a), the unshielded policy exhibits persistent low-amplitude oscillations and sporadic late-stage impulses in both kinetic-energy and velocity profiles. Adding a first–order low–pass filter (LPF) attenuates high-frequency motion but it is oblivious to the robot’s energetic limits, and exploration noise can still drive instantaneous commands beyond the admissible kinetic-energy envelope. 

Incorporating an eCBF shield mitigates jitter-induced motion by implementing an online, state-dependent projection that is invoked only when a commanded velocity would violate the bound $\tfrac12 m \lVert \mathbf v\rVert^{2} \le E_{\text{max}}$. The projection is direction-preserving and computed within a single control cycle, thereby imposing a hard safety invariant without compromising task performance.  Under identical stochastic sampling the shielded controller yields trajectories that converge monotonically and display no secondary transients as shown in Fig. 5(b). The combined LPF\,+\,eCBF architecture maintains the policy’s fastest admissible motions while provably suppressing the unsafe tails induced by stochastic control noise, a prerequisite for dependable real-world deployment.  
The eCBF shield complements the learned policy by enforcing hard safety bounds at runtime without retraining; however, it may scale down end-effector velocity when policy commands exceed kinetic energy limits. While this occasionally lengthens approach time, the effect is minimal compared to the safety benefits.

\begin{figure}
    \centering
    \vspace{10pt}
    \begin{subfigure}{1\linewidth}
        \centering
        \includegraphics[width=\linewidth]{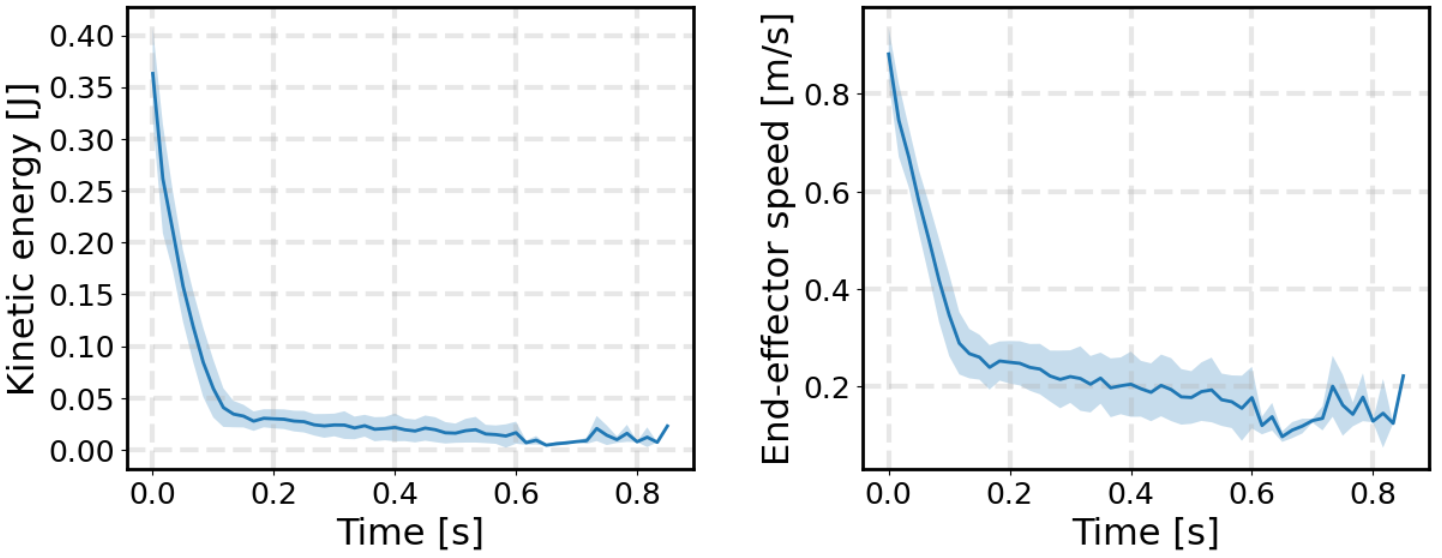}
        \caption{ContactRL without shield.}
        \label{fig:unshielded}
    \end{subfigure}
    
    \vspace{0.4em}  % vertical spacing between subfigures

    \begin{subfigure}{1\linewidth}
        \centering
        \includegraphics[width=\linewidth]{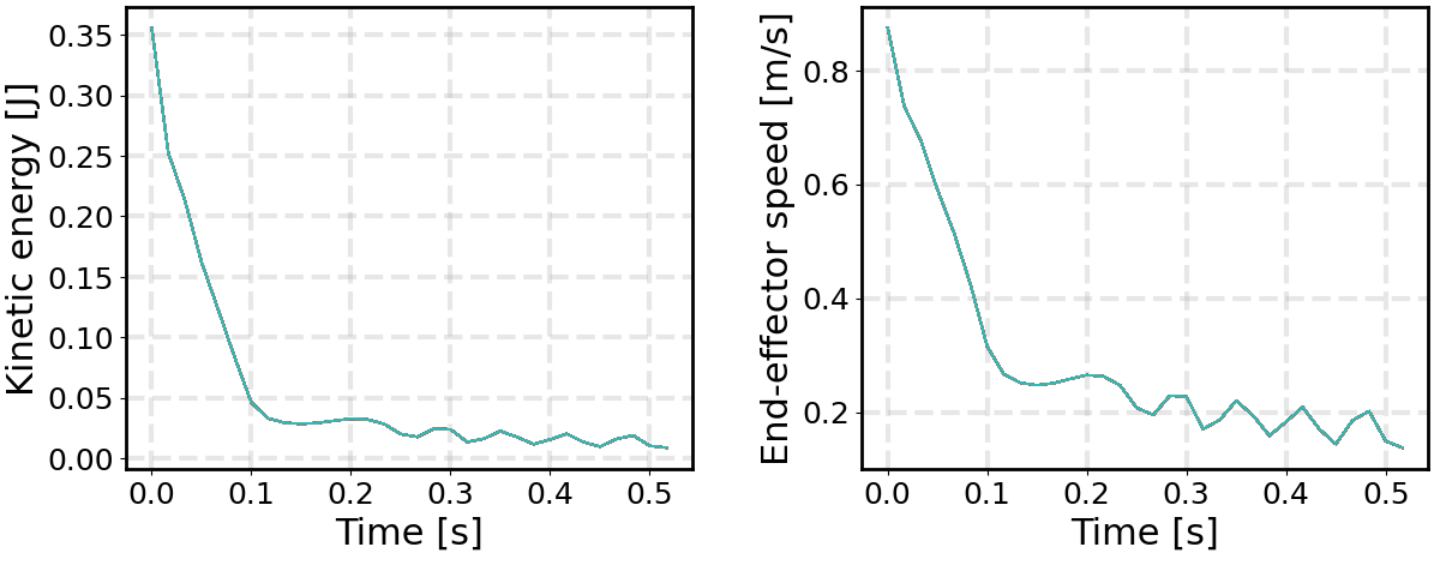}
        \caption{ContactRL with eCBF shield.}
        \label{fig:shielded}
    \end{subfigure}
    
    \caption{Kinetic energy (\textit{left}) and end-effector speed (\textit{right}) evaluated over 100 simulated episodes with stochastic action space. The eCBF shield results in smoother dissipation of kinetic energy and reduction of end-effector speed, enhancing stability and safety.}
    \label{fig:shielding_comparison}
    \vspace{-0.5em} % adjust value as needed
\end{figure}

\subsection{Contact Force Analysis with Physical Robotic Platform}
ContactRL policy with eCBF shield is deployed on the physical robotic platform, UR3e robotic arm equipped with a Robotiq 2F-85 gripper, to validate its performance and safety in real-world human to robot small object handover task. The UR3e’s integrated 6‑axis wrist force/torque (F/T) sensor, located at the flange just upstream of the tool‑centre point (TCP) measures forces. We recruited twelve healthy participants for our experiment. Five different objects were used to represent shape variability (Fig. 6). These objects were placed in participants’ palms at two different poses. For each object–pose combination, three trials were conducted. Raw six-axis force–torque data were captured at 250 Hz for every human–to–robot hand-over trial (N = 360, twelve participants × five objects × two poses × three trials). Each trial has different contact lengths due to different contact durations. For all trials, the force frame was aligned such that $F_z$ corresponded to the surface normal of the human palm. For each trial, the normal contact forces ($F_z$) were extracted and two scalar descriptors were computed: the maximum force, $F_{N,\text{max}}$, and the mean force, $\overline{F}_N$. These values were organized into a per-trial table with categorical factors: participant, object, pose, and replicate. A two-way repeated-measures ANOVA (within-subject factors: object and pose; subject: participant) was performed on $F_{N,\text{max}}$, since it reflects the worst-case load and thus represents a key safety indicator. 

\begin{figure}
    \vspace{-0.5cm}
    \centering
    \includegraphics[width=1\linewidth]{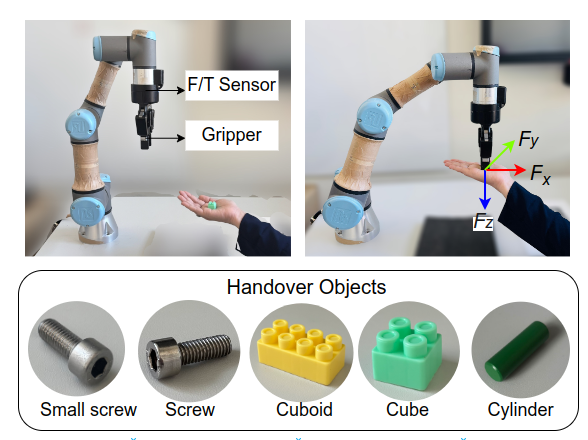}
    \caption{Experimental setup of the robotic platform. A UR3e robot, equipped with a force/torque (F/T) sensor and a parallel gripper, performs object handovers by grasping items directly from a participant’s hand while recording the interaction forces. The bottom row illustrates the objects used in the study.}
    \label{fig:exp-setup}
    \vspace{-0.5cm} % adjust value as needed
\end{figure}

\begin{figure*}
    \centering
    \vspace{10pt}
    \includegraphics[width=1\linewidth]{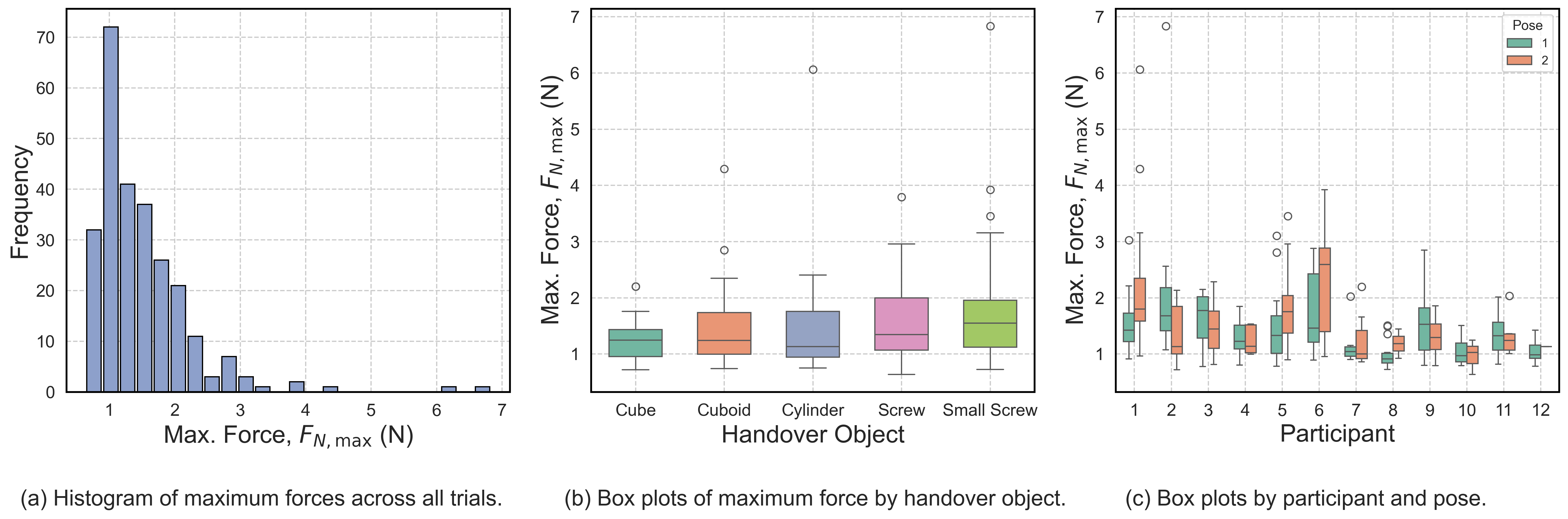}
    \caption{Contact force analysis of ContactRL with eCBF shield in real world small object handover trials. The results show that ContactRL policy with eCBF shield consistently achieves contact force, $F_N$, below 10 N indicating safe contact within human comfort limits.}
    \label{fig:force_analysis}
    \vspace{-0.5cm}
\end{figure*}

Fig. 7(a) shows a right-skewed distribution of maximum forces, with the majority of trials below $3$~N and a mode near $1$~N. Across all participants and conditions, $95\%$ of trials remained under $7$~N, confirming that the robot-human interactions stayed within the tactile comfort range for safe physical contact.  Object geometry significantly influenced maximum forces ($F_{4,28} = 23.1$, $p < 0.001$, $\eta^2 = 0.63$). As shown in Fig. 7(b), the \textit{cylinder} and \textit{screw} exhibited the highest medians, whereas the \textit{cube} consistently produced the lowest forces. Post-hoc tests confirmed that cylinder $>$ small screw ($p < 0.001$) and cylinder $>$ cube ($p = 0.008$). The effect of pose was also significant ($F_{1,7} = 12.4$, $p = 0.009$, $\eta^2 = 0.18$), with pose~2 (closer approach) yielding higher contact forces across all objects compared to pose~1 (far-side approach). No significant interaction between object and pose was found ($p = 0.41$), indicating that object-specific force trends were consistent across both poses. Inter-participant variability was notable. Fig. 7(c) illustrates individual differences in contact force regulation. While some participants consistently allowed higher robot-applied forces, others demonstrated greater caution, resulting in substantially lower force levels. These differences likely reflect variations in hand positioning and individual comfort thresholds during handover. Overall, the results verify that ContactRL with eCBF shield achieves safe handovers within human comfort limits, while highlighting systematic effects of object geometry, pose, and individual user behavior.

\subsection{Limitations of Sim-to-Real Transfer}
While our ContactRL policy with eCBF shield demonstrated safe contact force behavior on the physical platform, minor discrepancies were observed between simulated and real world trials. We evaluated the final position error across 30 real world reaching trials with randomized goal positions sampled within a bounded workspace defined by $x \in [-0.25, 0.25]~\text{m}$, $y \in [-0.30, -0.15]~\text{m}$, and $z \in [0.15, 0.40]~\text{m}$. The mean position error across trials was $0.0168 \pm 0.0188~\text{m}$ along the x-axis, $0.0269 \pm 0.0265~\text{m}$ along the y-axis, and $0.0420 \pm 0.0421~\text{m}$ along the z-axis (as shown in Table.III).
The final end-effector position errors were analyzed across all reaching trials to assess accuracy within a 1\,cm tolerance band. Approximately 56.7\% of trials achieved $|e_x| \leq 0.01\,\text{m}$, 73.3\% achieved $|e_y| \leq 0.01\,\text{m}$, and 53.3\% achieved $|e_z| \leq 0.01\,\text{m}$. Overall, 50\% of the reaches resulted in a radial position error below 1\,cm.

These deviations primarily stem from two factors. First, the sim-to-real transfer gap introduces modeling discrepancies between the simulated and physical robot dynamics. Differences in joint friction, actuator latency, and contact modeling can lead to slight deviations in the executed trajectories. Incorporating higher-fidelity physics engines or adaptive simulation calibration could improve correspondence between simulated and real behaviors. Second, the inherent safety–accuracy trade-off of the proposed framework limits positional precision. The enforcement of safety through eCBF-constrained control suppresses corrective motions near the target, resulting in marginal reductions in reaching accuracy. During the user studies, participants were instructed to act cautiously, and in some cases, they made small manual adjustments to complete the pickup task. These adjustments did not affect force measurements but helped compensate for the residual positioning error. Future work will focus on narrowing the sim-to-real gap by incorporating more realistic simulation physics and adaptive domain randomization techniques, enabling improved policy transfer while preserving the safety guarantees established by the eCBF shield.

\begin{table}
\centering
\caption{Mean, standard deviation, and percentage of 30 real world reaching trials within 0.01 m error tolerance for each Cartesian axis. The results show that the ContactRL policy with eCBF shield achieves centimeter scale accuracy.}
\begin{tabular}{lccc}
\toprule
Metric & Mean Error [m] & Std. [m] & $\%$ $\leq$ 0.01 [m] \\
\midrule
$|e_x|$ & 0.0168 & 0.0188 & 56.7 \%\\
$|e_y|$ & 0.0269 & 0.0265 & 73.3 \%\\
$|e_z|$ & 0.0420 & 0.0421 & 53.3 \%\\
\bottomrule
\end{tabular}
\label{tab:error_stats}
\vspace{-0.5cm}
\end{table}
\subsection{Discussion}
Variable impedance control (VIC) is often employed to regulate physical interaction by adjusting stiffness and damping online. However, VIC does not provide formal guarantees on maximum force or kinetic energy, and its performance is highly sensitive to human hand impedance, which varies across participants and trials. Achieving consistent behavior typically requires object or participant specific gain retuning, reducing reproducibility. In contrast, the proposed ContactRL with eCBF shield framework establishes an explicit safety envelope that is independent of hand impedance and object geometry. By decoupling performance optimization from safety enforcement, the learned policy maintains fluency while the shield intervenes when necessary, thereby avoiding the trade-off inherent in conservative VIC tuning. Our experimental results further highlight this advantage, showing consistent force bounds across objects, poses, and participants without the need for parameter adjustments.

\section{CONCLUSIONS}
Unlike prior safe RL frameworks that equate safety with non-contact, ContactRL explicitly enforces safe, intentional contact maintaining sub-10 N contact forces in all trials. The ContactRL framework offers two key advantages: (i) Safety is explicitly incorporated during training and enforced during execution, and (ii) The robot can plan safe object retrieval directly from the human palm, supporting secure handling of objects that cannot be reliably transferred in alternative poses. Beyond handovers, this approach has the potential to generalize to assistive scenarios requiring intentional physical contact, such as dressing and bathing. We demonstrated that ContactRL, augmented with an eCBF shield, ensures safety in contact-rich tasks through both simulation and real-world experiments. Future work will include extended safety assessment metrics and subjective evaluation as participant comfort also depends on factors such as impulse, contact duration, and contact area.

\section*{Ethical Approval}
All experimental procedures involving human participants were conducted in accordance with institutional ethical guidelines and approved by the appropriate ethics committee. Informed consent was obtained from all participants.

\bibliographystyle{IEEEtran}
\bibliography{references}
\end{document}